\documentclass{article}

\usepackage{PRIMEarxiv}

\usepackage[utf8]{inputenc} 
\usepackage[T1]{fontenc}    
\usepackage{hyperref}       
\usepackage{url}            
\usepackage{booktabs}       
\usepackage{amsfonts}       
\usepackage{nicefrac}       
\usepackage{microtype}      
\usepackage{lipsum}
\usepackage{graphicx}
\usepackage{amsmath}
\usepackage[numbers]{natbib}

\title{PatchFlow: Leveraging a Flow-Based Model with Patch Features\thanks{This work is supported by Stellantis}}

\author{
  Boxiang Zhang, Baijian Yang, Xiaoming Wang \\
  Purdue University \\
  West Lafayette, IN 47907, USA\\
  \texttt{\{zhan4653, byang, wang1747\}@purdue.edu} \\
   \And
  Corey Vian \\
  Stellantis - Kokomo Casting Plant \\
  Kokomo, IN 46902, USA \\
  \texttt{corey.vian@stellantis.com} \\
}

\begin{document}
\maketitle
\footnotetext{\protect
Author-prepared preprint accepted to the AAAI Workshop on AI for Autonomous and Embedded Systems (AI2ASE) 2024. Proceedings are non-archival.
}

\begin{abstract}
Die casting plays a crucial role across various industries due to its ability to craft intricate shapes with high precision and smooth surfaces. However, surface defects remain a major issue that impedes die casting quality control. Recently, computer vision techniques have been explored to automate and improve defect detection. In this work, we combine local neighbor-aware patch features with a normalizing flow model and bridge the gap between the generic pretrained feature extractor and industrial product images by introducing an adapter module to increase the efficiency and accuracy of automated anomaly detection. Compared to state-of-the-art methods \cite{roth2021total}, our approach reduces the error rate by 20\% on the MVTec AD dataset \cite{8954181}, achieving an image-level AUROC of 99.28\%. Our approach has also enhanced performance on the VisA dataset \cite{zou2022}, achieving an image-level AUROC of 96.48\%. Compared to the state-of-the-art models, this represents a 28.2\% reduction in error.  Additionally, experiments on a proprietary die casting dataset yield an accuracy of 95.77\% for anomaly detection, without requiring any anomalous samples for training. Defect segmentation on MVTecAD dataset is shown in Figure \ref{fig:localization}. Our method illustrates the potential of leveraging computer vision and deep learning techniques to advance inspection capabilities for the die casting industry.\end{abstract}  
\section{Introduction}
Die casting is a method where molten metal is injected under high pressure into a mold formed by two hardened steel dies. This process resembles injection molding. It's often used for high-volume production due to the significant investment required for the casting equipment and dies. Despite the high initial costs, the actual manufacturing of parts using die casting is relatively straightforward \cite{sai2017critical}. Die casting is crucial in modern manufacturing due to its precision, efficiency, versatility, and cost-effectiveness, especially for large-scale production. It plays a crucial role in various industries, including automotive, aerospace, and electronics. Despite these advantages, one of the primary issues that haunts the die casting industry is surface defects in cast products\cite{aivision}.

Surface Defects include blisters, peeling, or pits on the surface of the casting. They are often caused by contaminants in the metal, excessive moisture in the mold, or high metal temperature. Surface defects in die casting can lead to a variety of problems, including reduced product quality, increased production costs due to waste, and potentially detrimental impacts on product performance. In the worst-case scenario, these defects can lead to complete product failure. Detecting these flaws, especially in aluminum die casting, presents considerable challenges. Traditional inspection methods, like visual checks and other manual or semi-manual techniques, tend to be labor-intensive, time-consuming, and susceptible to human error. 

In light of these challenges, there is a growing need for an automated, efficient, and reliable method to detect surface defects in die casting products. Artificial Intelligence (AI), particularly anomaly detection in computer vision, emerges as a solution to address this issue\cite{aivision}.
\begin{figure}[ht]
\centering
\includegraphics[width=0.99\textwidth]{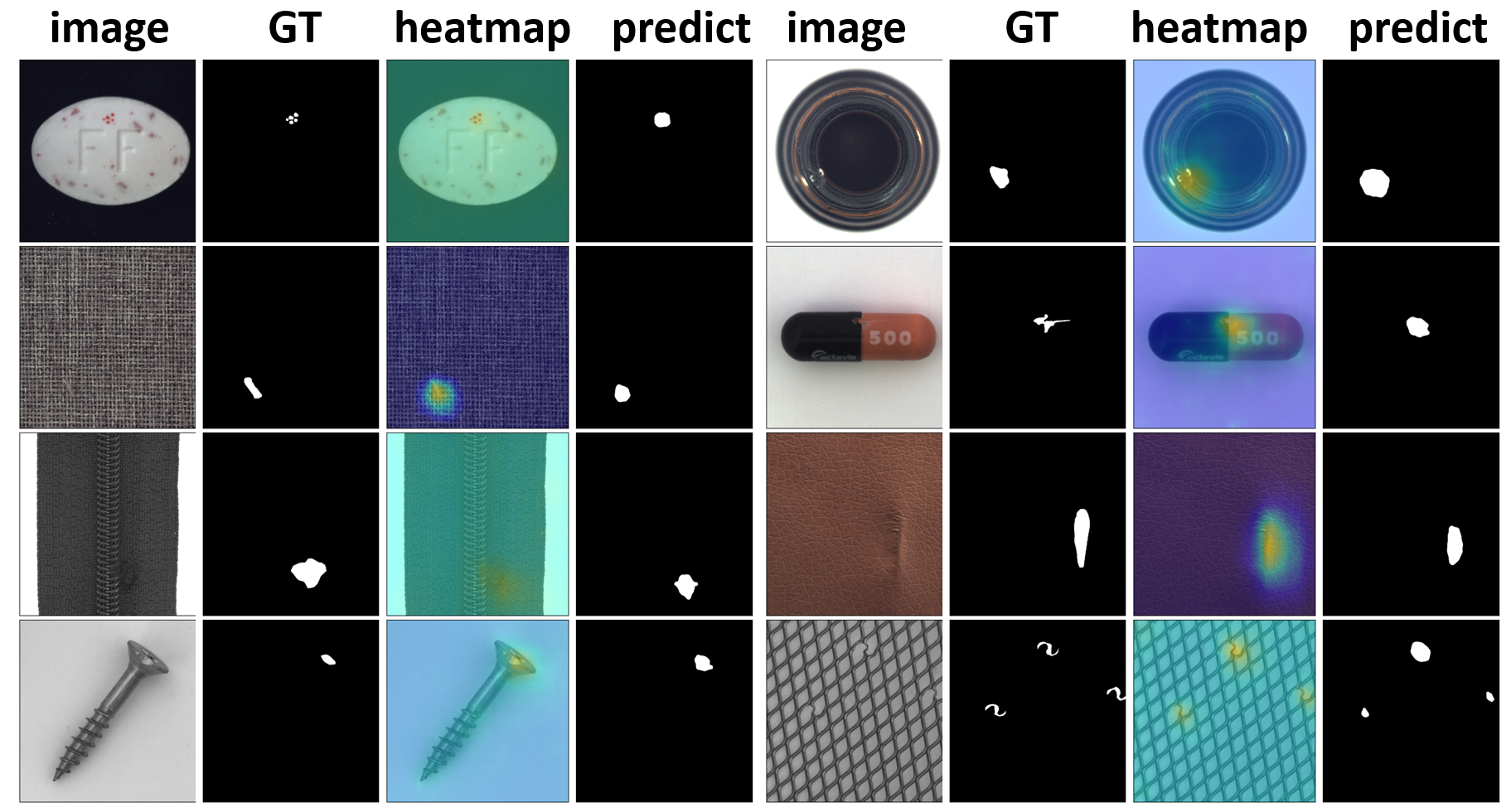} 
\caption{Visualization of per-pixel anomaly Groud Truth; heatmaps and predict anomaly mask}
\label{fig:localization}
\end{figure}
Anomaly detection is the process of identifying data points or patterns that deviate significantly from the expected behavior or norm, proves valuable in tackling quality control challenges in die casting. Figure \ref{fig:localization} displays eight examples of products with surface defects, illustrating the pixel-level ground truth of these defects, their corresponding heat maps, and the defects predicted using anomaly detection algorithms. 

Anomaly detection models can be viewed as semi-supervised learning models~\cite{8954181}. During the training phase, these models are trained exclusively on normal samples, without any exposure to anomalous instances. During the inference  stage, the model is presented with new instances and tasked with determining whether each instance is normal  or anomalous.The assumption is that the model will learn the underlying distribution of the normal data, and anything that deviates significantly from this learned distribution can be considered an anomaly. This decision is usually based on some measure of how much each instance deviates from the normal instances that the model was trained on. In computer vision, an instance would typically be an image or a video, and the measure of deviation could be the reconstruction error from a generative model, the feature representations learned by a deep neural network, or the likelihood under a probabilistic model.

Much research has focused on improving the accuracy and speed of vision-based anomaly detection tasks~\cite{roth2021total,zhang2023prototypical,lee2022cfa,Tien_2023_CVPR,yu2021fastflow, Gudovskiy_2022_WACV,ardizzone2019guided,RudWan2021,RudWeh2022}.
These studies have traditionally employed Convolutional Neural Network (CNN) architectures, such as ResNet~\cite{he2015deep} and EfficientNet~\cite{tan2020efficientnet}, to extract features for anomaly detection.

Embedding-based models require a memory bank, whereas flow-based models necessitate a sophisticated structure with numerous parameters, leading to a high computational load (FLOPs) for mapping the feature distribution to a latent normalized distribution. This mapping is essential for accurately calculating the exact distribution of features in normal samples within the training set. The objective of this work is to streamline the model and enhance the efficiency of flow-based architectures.

\textbf{The main contribution of this work is list below}

\begin{itemize}
    \item We propose \textbf{PatchFlow}, a patch-level anomaly detection framework that combines local neighborhood-aware features with flow-based likelihood modeling.
    \item We introduce a lightweight feature adaptation module that aligns pretrained representations with industrial image distributions, improving flow stability and detection accuracy.
    \item We design an efficient bottlenecked coupling structure for normalizing flows, reducing computational complexity while maintaining expressive capacity.
    \item We demonstrate state-of-the-art image-level anomaly detection performance on MVTec AD and VisA, and validate the approach on a real industrial die casting dataset.
\end{itemize}

The paper is organized as follows: The literature review is presented in the Related Works section, followed by an illustration of the proposed PatchFlow in the Method section. The Experiment section details the dataset, evaluation metrics, and AUROC results. Finally, a brief discussion on future work is covered in the Conclusion section. 

\section{Related Works}

The primary objective of anomaly detection in computer vision is to identify deviations from the norm within visual data. This process involves differentiating between standard, expected patterns and those that are irregular or unusual, which are referred to as anomalies. These anomalies can range from minor variations to significant deviations and are crucial for applications such as quality control in manufacturing, surveillance, and medical imaging. Anomaly detection models in computer visions can be broadly classified into three categories: generative-based models, feature-based models, and normalizing flow-based models.

\subsection{Generative-based Model}
Generative-based models are a class of machine learning models that focus on generating or simulating data. 
The underlying assumption is that these models can learn the data distribution of normal instances. In anomaly detection, generative models are typically trained exclusively on normal data. Once trained, the models attempt to reconstruct new input data based on what they have learned. If the new input data is normal, the models should be able to reconstruct it accurately. 
However, if the input data contains anomalies, the models will likely struggle to reconstruct it accurately, resulting in a high reconstruction error, which is the basis for anomaly detection.

Generative Adversarial Networks (GANs)~\cite{goodfellow2014generative},UNet~\cite{ronneberger2015unet}, Variational Autoencoders (VAEs)~\cite{kingma2022autoencoding} and diffusion models~\cite{sohldickstein2015deep}, are capable of generating new data instances that resemble the normal instances in the training set.

These models are often complex and require substantial computational resources. Scaling these models for large datasets or high-dimensional data can be challenging. In addition, Generative models can be difficult to train. For example, GANs involve training a generator and a discriminator simultaneously. Models can easily become unstable or fail to converge.
\subsection{Feature-based Model}
In traditional Feature-Based Models, features are explicitly defined and engineered based on domain knowledge. For example, in image processing, features might include edge detection, color histograms, or texture measures. These features are manually crafted and selected to capture relevant information for a specific task. The performance heavily relies on the quality of feature engineering, which can be labor-intensive and may not capture all the nuances in complex data.

Embedding-based models are a modern evolution of feature-based models, primarily used in the context of deep learning. Instead of manually defining the features, the model learns an optimal representation of the data autonomously. These learned representations, known as embeddings, are essentially high-dimensional feature vectors. 
Embeddings capture complex and abstract patterns in the data, which might not be easily identifiable with traditional feature engineering.
Embedding-based models have become increasingly popular in anomaly detection. These models learn a compact representation of normal data and then identify deviations from this norm as potential anomalies. 
Various techniques can be utilized for feature extraction, spanning simple image statistics to complex deep learning architectures. PADiM~\cite{DBLP:journals/corr/abs-2011-08785} introduced a feature aggregation technique to improve model performance by combining multi-level descriptors extracted from different layers of a CNN. Specifically, features are computed from both shallow and deep layers, encapsulating low-level visual patterns along with higher-level semantic information.  PatchCore~\cite{roth2021total} employs WideResnet50\cite{zagoruyko2017wide} as the backbone feature extractor, combined with a feature aggregation module that incorporates neighboring descriptors to enrich the representations. Anomaly detection is performed by comparing each instance's embedding against a memory bank of features computed from normal samples. Deviations from the distribution of normal embeddings indicate the presence of defects. Thus, embedding-based approaches rely on learning a robust feature space where anomalies separate clearly from defect-free patterns. 
\\
\subsection{Normalizing Flow-based Model}
Normalizing Flows (NF)~\cite{rezende2016variational} are a class of neural networks that can learn bijectivity transformation $N$ between data distribution $a\in A$ and well-defined densities $b\in B$, such that $B=N(A)$ and $A = N^{-1}(B)$. The well defined distribution could be the standard normal distribution such that $B \sim \mathcal{N}(0,I)$. According to the Change of Variable Theorem, the relation of the distribution of $A$ and $B$ can be defined as:
\begin{equation}
\hat{p_A}(a) =p_B(N(a))\lvert \text{det}(\frac{\partial N(a)}{\partial a^T}) \rvert
\label{eq:pA}
\end{equation}

A key advantage of NF models is that the transformation $N:A\to B$ enables the calculation of the exact distribution of $A$  by equation \ref{eq:pA}. Real-NVP \cite{dinh2017density} introduces a specific type of normalizing flow that not only uses easily invertible transformations but also has a Jacobian determinant that is straightforward to compute.

In the field of anomaly detection, normalizing flow-based models can learn the exact probability density of the training data and provide a way to transform a simple distribution (like a multivariate Gaussian) into a complex one that matches the distribution of the data. Anomalies can be detected based on their low probability under this learned distribution. In this work we employ normalizing flow-based models to detect anomaly.

DifferNet~\cite{RudWan2021} innovatively  integrate AlexNet~\cite{10.1145/3065386} and a normalizing flow structure with RealNVP~\cite{dinh2017density} coupling layer for the task of anomaly detection. It demonstrated impressive capabilities in terms of image-level prediction accuracy, even without the requirement for a large volume of training data. However, despite its achievements, it did not deliver high performance at the pixel level, signifying an area for potential improvement.
Subsequent work, CsFlow~\cite{RudWeh2022},  increased the input resolution of the feature extractor and using a more powerful feature encoder EfficientNet~\cite{tan2020efficientnet} as well as introduced the concept of a cross-scale flow structure. This structure facilitated interaction of features from different layers within the normalizing flow . This resulted in a performance enhancement over DifferNet.
CFLOW-AD~\cite{Gudovskiy_2022_WACV}  enhanced the application of normalizing flow in anomaly detection by introducing conditional normalizing flow~\cite{ardizzone2019guided} with positional encoding, which provides position prior information. 
FastFlow~\cite{yu2021fastflow} integrated transformers with normalizing flow-based anomaly detection models by substituting the feature extractor with Class-Attention in Image Transformers (CaiT)~\cite{touvron2021training} and Data-efficient Image Transformers (DeiT)~\cite{touvron2021going}.

All these models utilize features directly extracted from the pretrained backbone. However, the discrepancy of distribution between the images on which the backbone model is pretrained and the industrial images poses a constraint on detection accuracy. Additionally, normalizing flow-based models typically entail a complex structure for distribution mapping, resulting in high computational costs and slow inference speeds. In this study, we mainly focus on addressing the gap of normalizing flow based model.

\section{Method}
\subsection{Model Overview}
\begin{figure*}[ht]
\centering
\includegraphics[width=0.99\textwidth]{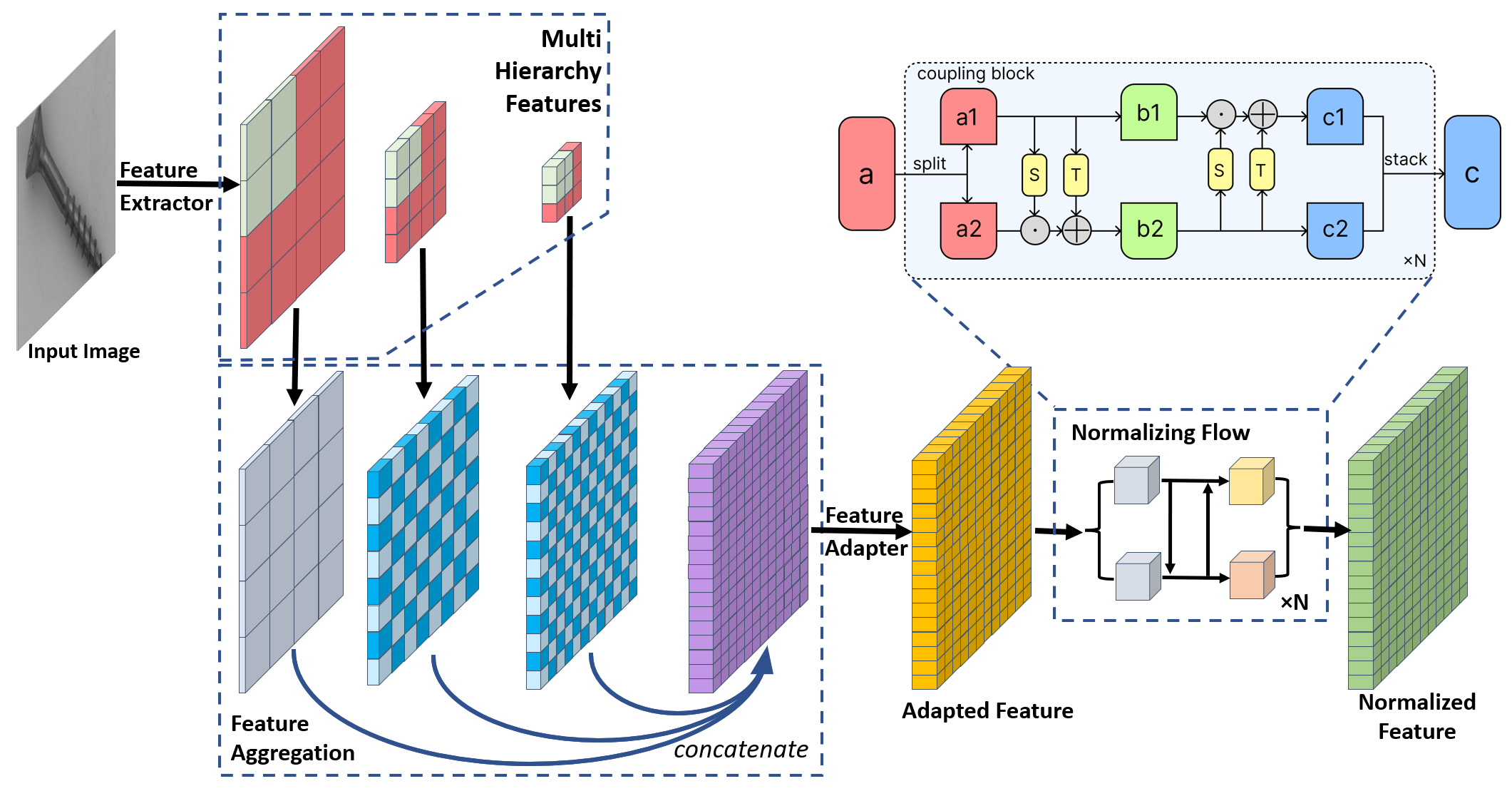} 
\caption{Model Overview. The PatchFlow model comprises four key components. 1, a pretrained feature extractor extracts multi-level representations from multi-scale images. 2, a feature aggregation layer combines descriptors from different hierarchical levels and scales. 3, a feature adapter module reduces the dimensionality of the representations and bridges the gap between the generic pretraining data and specialized industrial product images. 4, a normalizing flow maps the adapted features to a standardized distribution.  }
\label{fig:overview}
\end{figure*}

PatchFlow consists of four main components, illustrated in Figure~\ref{fig:overview}.
First, a pretrained feature extractor produces multi-level representations from input images.
Second, a feature aggregation module constructs local neighborhood-aware patch features across multiple hierarchies and scales.
Third, a feature adaptation module reduces feature dimensionality and mitigates distribution mismatch between pretrained features and industrial data.
Finally, a normalizing flow maps the adapted patch features to a standard normal distribution, enabling likelihood-based anomaly detection.
This design allows PatchFlow to model patch-level feature distributions efficiently, supporting both image-level anomaly scoring and pixel-level defect localization.

\subsection{Feature Extractor}
Feature extractor $\mathcal{E}$ maps data $x\in \mathbb{R}^{ C\times H \times W}$  with $C$ channels, height $H$ and width $W$ from dataset $X=\{x_1,...,x_D\}$ of size $D$ into a latent space $y \in Y$. For each $i^{th}$ data point, this relationship is denoted as $y_i = \mathcal{E}(x_i)$. To ensure the preservation of both the global abstract information and local special information inherent in the images, instead of solely relying on the features from the final layer, we utilize features from the multiple layers of the feature extractor. $y_i^j = [E(x_i)]^j$ refers to the feature from hierarchy $j\in\{1,2,...\}$of the feature extractor for the image $i\in\{1,2,...,D\}$. The feature extractor is well trained on large dataset like ImageNet\cite{deng2009imagenet} and keep fix for both training and testing.\\

\subsection{Feature Aggregation}
After obtaining features from a feature extractor across different hierarchies, we employed a feature aggregation layer $\mathcal{P}$ to combine the local neighborhood, which consisted of a feature patcher and a featur fuser, both of which do not introduce new trainable parameters. 
For the patcher, following the approach outlined in \cite{roth2021total}, it combine local neighborhood for each hierarchy respectively. Let $y_i^j \in \mathbb{R}^{c^j, h^j, w^j}$, where $c^j$, $h^j$, and $w^j$ represent the dimensions, height, and width of the feature from hierarchy $j$. 
With CNN structure, typically, $h^{j}>h^{j+1}$, $w^{j}>w^{j+1}$, $c^{j}<c^{j+1}$ . 
This is extended to $y_i^j(h, w) \in \mathbb{R}^{c^j}$, indicating the feature slice at position $(h, w)$, where $h \in \{1, \ldots, h^j\}$ and $w \in \{1, \ldots, w^j\}$. 
The neighborhood of position $(h,w)$ with uneven patch size $\mathcal{s}$ is defined as follows: 
\begin{equation}
\begin{split}
N_\mathcal{s}^{(h,w)} = \{(a,b)|&a \in [h-[\mathcal{s}/2],..,h+[\mathcal{s}/2]],\\
& b\in [w-[\mathcal{s}/2],...,w+[\mathcal{s}/2]]\}
\end{split}
\label{eq:neighbor}
\end{equation}
The local neighbourhood aware patch feature of image $i$ from hierarchy $j$ at position $(h,w)$ with patch size $\mathcal{s}$ is defined as: 
\begin{equation}
p^j_i(N_\mathcal{s}^{(h,w)})=Agg(y_i^j(a,b)|(a,b) \in N_\mathcal{s}^{(h,w)})
\label{eq:agg}
\end{equation}
where $Agg$ is some aggregation function. 
After obtaining patch features from each hierarchy,  $p^j_{i,\mathcal{s}}=Agg(y_i^j)_\mathcal{s}\in \mathbb{R}^{c^j,h^j,w^j}$, with $p^j_{i,\mathcal{s}}(h,w)=p^j_i(N_\mathcal{s}^{(h,w)})\in \mathbb{R}^{c^j}$. These features are fused together to create a multi-hierarchy local aware feature. This is achieved by interpolating the deeper features to a size that matches the features from the shallower layers and concatenating them together, as illustrated in the figure. 
The feature aggregation layer $\mathcal{P}$ takes features from different hierarchies as input and outputs a fused local aware patch feature,  $f_{i,\mathcal{s}}=\mathcal{P}_\mathcal{s}(y_i^1,y_i^2,...)$, with $f_{i,\mathcal{s}}(h,w)\in \mathbb{R}^{C_{f}}$, where $C_{f}=\sum_j c^j$. 
In addition to incorporating features from multiple hierarchies, our approach involves leveraging features at different scales. Features from various scales undergo the same processing pipeline, yielding several patch features of the same size. These patch features are then straightforwardly concatenated to form a multi-scale hierarchy patch feature. In our work, we specifically use three scales. Consequently, the multi-scale hierarchy patch feature for data $i$ with patch size $\mathcal{s}$ can be expressed as $f^3_{i,\mathcal{s}}(h,w)\in \mathbb{R}^{3\times C_{f}}$.
\begin{figure*}[ht]
\centering
\includegraphics[width=0.99\textwidth]{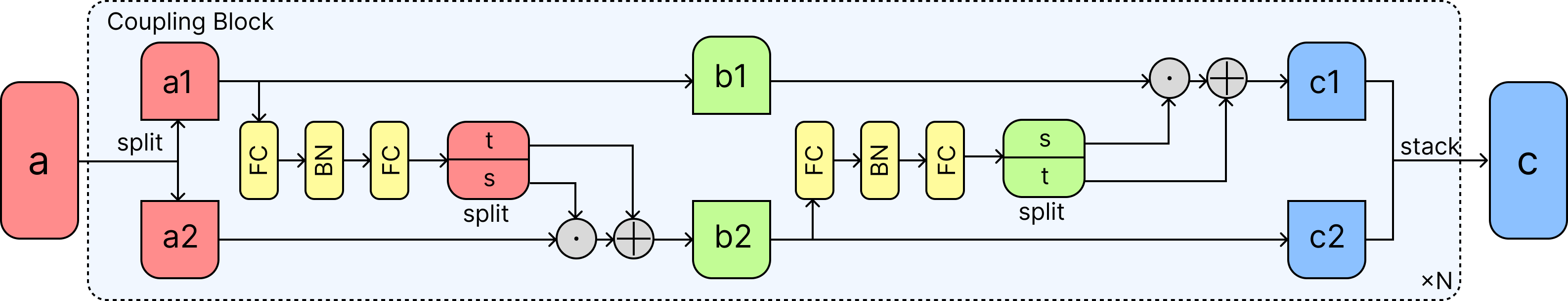} 
\caption{The coupling block features a bottleneck structure, where 'FC' and 'BN' refer to the fully connected layer and bottleneck layer.}
\label{fig:coupling_block}
\end{figure*}
\subsection{Feature Adaptor} 
The industrial dataset exhibits a distinct distribution compared to the dataset on which the feature extractor was originally trained. 
To mitigate the impact of these distributional differences and improve the model's performance on the industrial data, we introduce an adaptation layer denoted as $\mathcal{A}$. This layer facilitates the transfer of patch features $f_{i,\mathcal{s}}(h,w)$ to adapted features $g_{i,\mathcal{s}}(h,w)\in\mathbb{R}^{C_{g}}$, to address these distributional differences and enhance the model's effectiveness in handling the industrial data. 
\begin{equation}
g_{i,\mathcal{s}}(h,w) = \mathcal{A}(f_{i,\mathcal{s}}(h,w))
\label{eq:agg_1}
\end{equation}
To achieve this adaptation, we employ a single fully-connected layer, a strategy demonstrated to be effective in prior work such as \cite{Liu_2023_CVPR}. By incorporating this adaptation mechanism, we aim to enhance the model's adaptability to the specific distributional nuances encountered in the industrial setting.

\subsection{Flow}
To transform the distribution $P_G(g(h,w))$ of the adapted feature $g(h,w)$, features are subsequently passed through a bijective mapping $\mathcal{F}$, where  $\mathcal{F}: G \to Z$. The aim here is to bijectively map the features $g(h,w)\in G$ to a normalized latent space $z(h,w) \in Z$ with normal distribution, such that $Z \sim \mathcal{N}(0,I)$.
This bijective transformation is represented as $z(h,w) = \mathcal{F}(g_{s}(h,w)$,  $g_{s}(h,w) = \mathcal{F}^{-1}(z(h,w))$. The estimate distribution $\hat{g}(x,y)$ of $g(x,y)$ is given by equation\ref{eq:pA}
In contrast to prior approaches \cite{ardizzone2019guided,Gudovskiy_2022_WACV,RudWan2021,yu2021fastflow} that map the distribution of image-level features to a normalized latent distribution, our method directly maps each patch-level feature. This strategy offers the advantage of facilitating straightforward defect localization based on the position of normalized latent patch features that deviate from the normal distribution. This approach allows for a more fine-grained and precise identification of defects, as the deviation from the normal distribution is localized at the patch level rather than the entire image.

For each Bijective Mapping block, we follow previous works~\cite{ardizzone2019guided,Gudovskiy_2022_WACV,RudWan2021,yu2021fastflow}, which adopt the coupling layer structure from RealNVP~\cite{dinh2017density} as the basic building block in our PatchFlow architecture (see Figure \ref{fig:coupling_block}). 

We adopted a fully connected layer with a bottleneck structure to preserve the capacity for mapping between distributions while reducing the computational complexity (Flops) in the coupling layer. This design choice allows us get a deeper network with less Flops~\cite{he2015deep}, while the bottleneck structure enhances efficiency by constraining the number of parameters involved in the transformation.  In our implementation, we employed a three-layer convolutional neural network with a kernel size of 1. Figure \ref{fig:coupling_block} shows the detail structure of coupling block.\\
\begin{table*}[ht]
\centering
\begin{tabular}{llllllll}
\toprule[2pt]
Model      & DRÆM & CutPaste & CFlow-AD   & CS-Flow & PADIM& PatchCore & \textbf{PatchFlow}(Ours) \\ 
\midrule[1pt]
Carpet     & 97.0 & 93.9   & 98.98 & \textbf{100}   & 99.1 & 98.7 &    \textbf{100}       \\ 
Grid       & 99.9 & \textbf{100}    & 97.64 & 99.0  & 97.3 & 98.2 &    99.83       \\ 
Leather    & \textbf{100}  & \textbf{100}    & 98.98 & \textbf{100}   & 99.2 & \textbf{100}  &    \textbf{100}       \\ 
Tile       & 99.6 & 94.6   & 99.25 & \textbf{100}   & 94.1 & 98.7 &    99.10       \\ 
Wood       & 99.1 & 99.1   & 98.99 & \textbf{100}   & 94.9 & 99.2 &    99.65       \\ 
Bottle     & 99.2 & 98.2   & 98.89 & 99.8  & 98.3 & \textbf{100}  &    \textbf{100}       \\ 
Cable      & 91.8 & 81.2   & \textbf{99.66} & 99.1  & 96.7 & 99.5 &    99.01       \\ 
Capsule    & 98.5 & 98.2   & \textbf{98.56} & 97.1  & 98.5 & 98.1 &    97.21       \\ 
Hazelnut   & \textbf{100}  & 98.3   & 98.95 & 99.6  & 98.2 & \textbf{100}  &    \textbf{100}       \\ 
Metal Nut  & 98.7 & 99.9   & 98.86 & 99.1  & 97.2 & \textbf{100}  &    \textbf{100}       \\ 
Pill       & \textbf{98.9} & 94.9   & 98.01 & 98.6  & 95.7 & 96.6 &    97.35       \\ 
Screw      & 93.9 & 88.7   & \textbf{98.93} & 97.6  & 98.5 & 98.1 &    98.34       \\ 
Toothbrush & \textbf{100}  & 99.4   & 97.99 & 91.9  & 98.8 & \textbf{100}  &    \textbf{100}       \\ 
Transistor & 93.1 & 96.1   & 96.65 & 99.3  & 97.5 & \textbf{100}  &    99.54       \\ 
Zipper     & \textbf{100}  & 99.9   & 99.08 & 99.7  & 98.5 & 99.4 &    99.21       \\ 
\midrule[1pt]
Average    & 98.03 & 96.1   & 98.62 & 98.7  & 97.5 & 99.1 &   \textbf{99.28}        \\ 
\bottomrule[2pt]
\end{tabular}
\caption{Image level AUROC score of DRÆM \cite{Zavrtanik_2021_ICCV}, CutPaste\cite{li_cutpaste_2021}, PADIM\cite{DBLP:journals/corr/abs-2011-08785}, PatchCore\cite{roth2021total}, CFlow-AD\cite{Gudovskiy_2022_WACV}, CS-Flow\cite{RudWeh2022} from the MVTecAD\cite{8954181} dataset are presented, with the best scores highlighted in bold.}
\label{table_result}
\end{table*}

\subsection{Objective Function}
The objective of the training process is to find optimized parameters $\theta$ within the normalizing flow $N$ to minimize the Kullback-Leibler (KL) divergence between the estimated feature distribution $\hat{f}(h,w)\in F$and the actual feature distribution $y\in Y$. We define a transformation $N: Y \to Z$ where $Z$ follows a standard normal distribution, $Z \sim \mathcal{N}(0,I)$. the KL divergence can be mathematically expressed as follows:
\begin{equation}
\label{eq:loss}
\begin{split}
    \mathcal{L}(\theta) & = D_{KL}^\theta(p_y|\hat{p_y}) \\
     & =\sum\limits_{i=1}^Dp_y(y_i)log(\frac{p_y(y_i)}{\hat{p_y}(y_i)}) \\
     & = \frac{1}{D}\sum\limits_{i=1}^D\frac{|| N(y_i)||_2^2}{2}-log(\lvert \text{det}(\frac{\partial N(y_i)}{\partial y_i^T}) \rvert) +c
\end{split}
\end{equation}
In equation \ref{eq:loss}, the constant term ($const$) is not directly linked to the model parameters $\theta$. Thus, when applying the back-propagation algorithm to optimize the loss function, this $c$ term does not influence the update of the model parameters. It is effectively treated as a constant component within the loss function, which simplifies the computation during the optimization process.

\subsection{Mask Generation and Score Function}
$z(h,w) \in \mathbb{R}^{C_{g}}$ such that $Z \sim \mathcal{N}(0,I)$. 
each element of the latent feature encapsulates spatial information. Therefore, any deviation from the normal distribution indicates not only an anomaly but also suggests its relative position.
In order to maintain simplicity in prediction, the generation of anomaly mask and the anomaly score function involve straightforward linear transformations of the normalized latent patch features. Given the normalized features $ z\in \mathbb{R}^{C_{g}\times F_g\times W_g}$, where  $C_g,H_g,W_g$ represent the number of channels, height, and width of the feature after feature adaption layer respectively. We upscale features to match the input image size using bilinear interpolation to generate the final anomaly map with size $R^{B\times H \times W}$. This map presents a comprehensive view of the anomalies detected in the input image, effectively leveraging information from various hierarchy and scales to enhance anomaly detection. The anomaly score for each image is then determined by the maximum value in the corresponding anomaly map.



\section{Experiments}
\subsection{Datasets}
Numerous datasets have been specifically designed for anomaly detection tasks. In this study, we evaluate our model using two well-established, large-scale anomaly detection datasets, namely the MVTec AD~\cite{8954181} dataset and VisA dataset\cite{zou2022}.\\
The MVTec AD~\cite{8954181} dataset is a comprehensive benchmark for anomaly detection methods. It's widely recognized in the field, comprising over 5000 high-resolution images categorized into 15 different object and texture types.\\

The VisA~\cite{zou2022} dataset consists of 12 subsets, each corresponding to a different object. It includes a total of 10,821 images, with 9,621 normal and 1,200 anomalous samples. The subsets range from complex structures such as different types of printed circuit boards (PCBs), to multiple instances within a single view, and objects that are mostly aligned. Anomalous images in this dataset present various flaws, including surface defects like scratches, dents, and color spots, and structural defects like misplacement or missing parts.

\begin{figure}[hb]
\centering
\includegraphics[width=0.5\columnwidth]{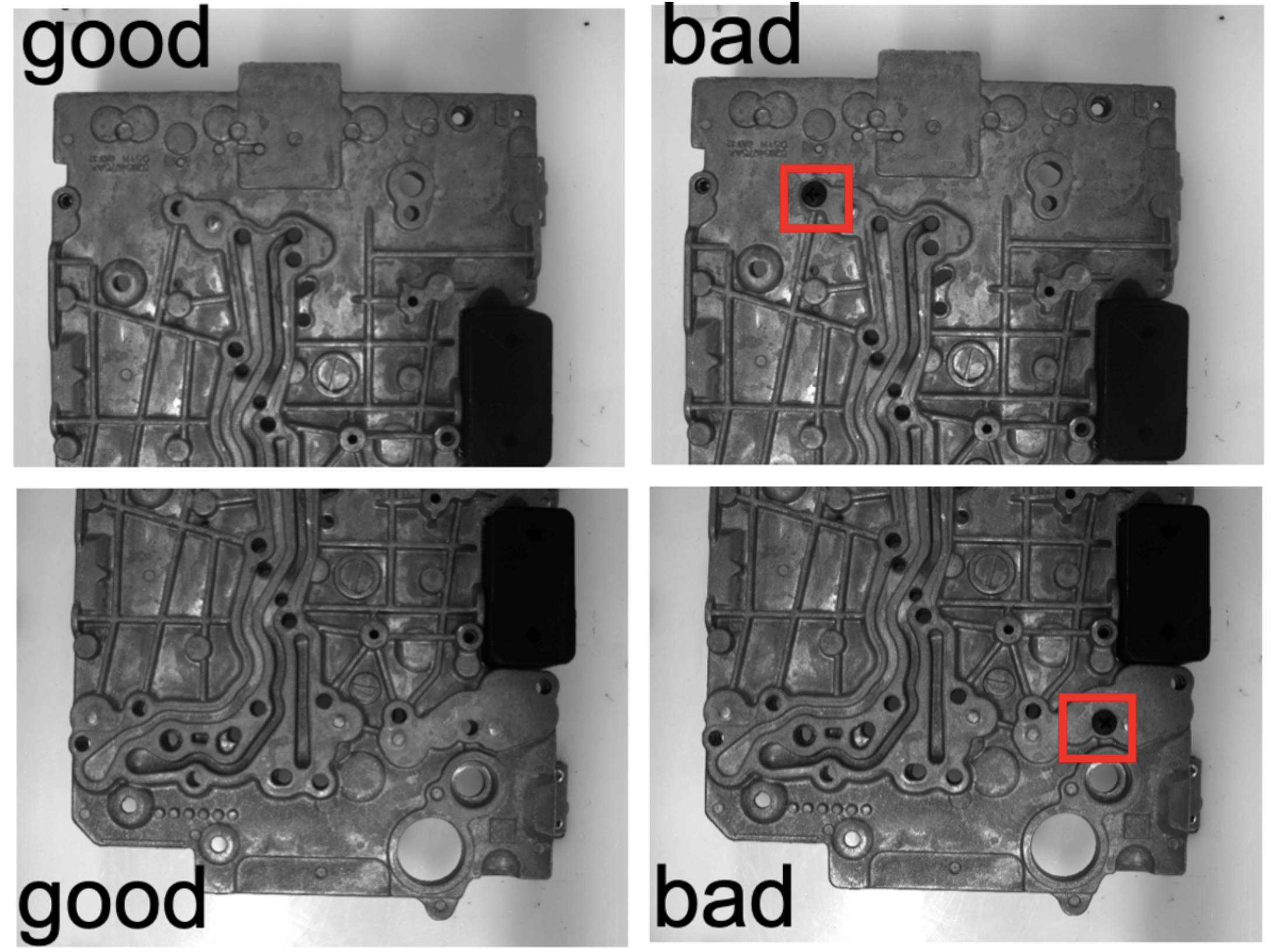} 
\caption{Demonstration of collected valve body data. The valve body data used for demonstration was collected with synthesized defects randomly generated by placing nails on the valve surface. }
\label{fig:valve_body_defect}
\end{figure}

We also collected images of valve bodies, each with a $2592\times1944$ resolution. In total we collected 112 defect-free samples and 36 synthesized anomaly samples generated by randomly placing nails on the valves as shown in Figure \ref{fig:valve_body_defect}. For testing, we used 34 of the 112 defect-free samples along with all 36 synthesized anomaly samples. The remaining 78 defect-free samples were used for training.

\begin{table*}[ht]
\centering
\begin{tabular}{ccccccc}
\toprule[2pt]
Models   & DRÆM & CutPaste & CFlow-AD & PADIM & PatchCore & PatchFlow \\ 
\midrule[1pt]
Im-AUROC & 88.7 & 81.9     & 91.5     & 89.1  & 95.1      & \textbf{96.48 }    \\ 
\bottomrule[2pt]
\end{tabular}
\caption{Image level AUROC score of DRÆM \cite{Zavrtanik_2021_ICCV}, CutPaste\cite{li_cutpaste_2021}, PADIM\cite{DBLP:journals/corr/abs-2011-08785}, PatchCore\cite{roth2021total}, CFlow-AD\cite{Gudovskiy_2022_WACV}, from the VisA\cite{zou2022} dataset are presented, with the best scores highlighted in bold.}
\label{table:VisA_result}
\end{table*}
\subsection{Evaluation Metrics}
Given that the anomaly detection model is exclusively trained on normal data, it can only produce anomaly scores at either the image or pixel level. To identify defects, a threshold is essential. To avoid manual threshold selection, following established methodologies, we calculate both image-level AUROC (Area Under the Receiver Operating Characteristic curve) and pixel-level AUPRO (Area Under the Precision-Recall curve) \cite{8954181} to evaluate the performance of our model in image-level classification and pixel-level defect localization.
\subsection{Implementation Details}
We resize the input images to $768 \times 768$. Importantly, our approach avoids the use of any data augmentation or normalization techniques on the input images. Features are extracted from the 12th, 19th, and 35th blocks of EfficientNet B5, forming a 3-scale feature pyramid with dimensions $1488 \times 96 \times 96$. Subsequently, a feature adaptor is employed to decrease the feature size to $768 \times 96 \times 96$. The computational complexity and FLOPs of our model scale proportionally with the dimensions of this reduced feature representation. The bottleneck dimension of the flow is fixed at 128.

\subsection{Results}
We compare PatchFlow with several state-of-the-art methods using different learning approaches, including: the generative model-based methods DRÆM \cite{Zavrtanik_2021_ICCV}, CutPaste\cite{li_cutpaste_2021}; the embedding-based model PADIM\cite{DBLP:journals/corr/abs-2011-08785}, PatchCore\cite{roth2021total}; and the flow-based model CFlow-AD\cite{Gudovskiy_2022_WACV}, CS-Flow\cite{RudWeh2022} on both MVTecAD dataset\cite{8954181} and VisA dataset\cite{zou2022}. The numerical results for image level anomaly detection on the MVTecAD dataset are listed in \ref{table_result}. As shown, the PatchFlow method outperforms the state-of-the-art PatchCore by reducing the error from 0.9\% to 0.72\%, which leads to a \textbf{20\% reduction of error} in terms of image-level anomaly detection, as measured by the image-level AUROC evaluation metric result in 99.28 auroc. PatchFlow consistently demonstrates high performance across various materials, making it a standout model in this evaluation. In table \ref{table:VisA_result} we compare the performance of various state-of-the-art models on VisA\cite{zou2022} dataset. PatchFlow reduce the error of 4.9\% to 3.52\% which leads to \textbf{28.2\% reduction of error} compared with PatchCore model. 

The improvement of PatchFlow over other models on both dataset is notably significant. The high AUROC score of PatchFlow indicates its robustness and reliability in various scenarios, making it a potentially more versatile and effective tool in practical applications. 

In the visualization presented in Figure \ref{fig:localization}, both per-pixel anomaly ground truth maps and corresponding predicted heatmaps and predicted anomalous mask are displayed. This illustration effectively showcases PatchFlow's capability, particularly highlighting its proficiency in accurately locating small defects. The detailed comparison between the ground truth and the model’s predictions underlines the precision with which PatchFlow can identify even minute anomalies.

In addition to assessing detection and localization performance, we conduct an evaluation of the computational complexity. The total FLOPs amount to $1.25\times10^{10}$, resulting in a processing speed of 2.7 frames per second (FPS) when utilizing a single Nvidia GeForce RTX2080Ti GPU.
\begin{table}[h]
\centering
\begin{tabular}{lcc}
\toprule[2pt]
   & Actual Normal & Actual Abnormal \\ 
   \midrule[1pt]
Predicted Normal & 34 & 3  \\ \hline
Predicted Abnormal  & 0  & 33 \\ 
\bottomrule[2pt]
\end{tabular}
\caption{confusion matrix of model prediction on collected valve body dataset.}
\label{table:confusion_matrix}
\end{table}

\begin{figure}[t]
\centering
\includegraphics[width=0.5\columnwidth]{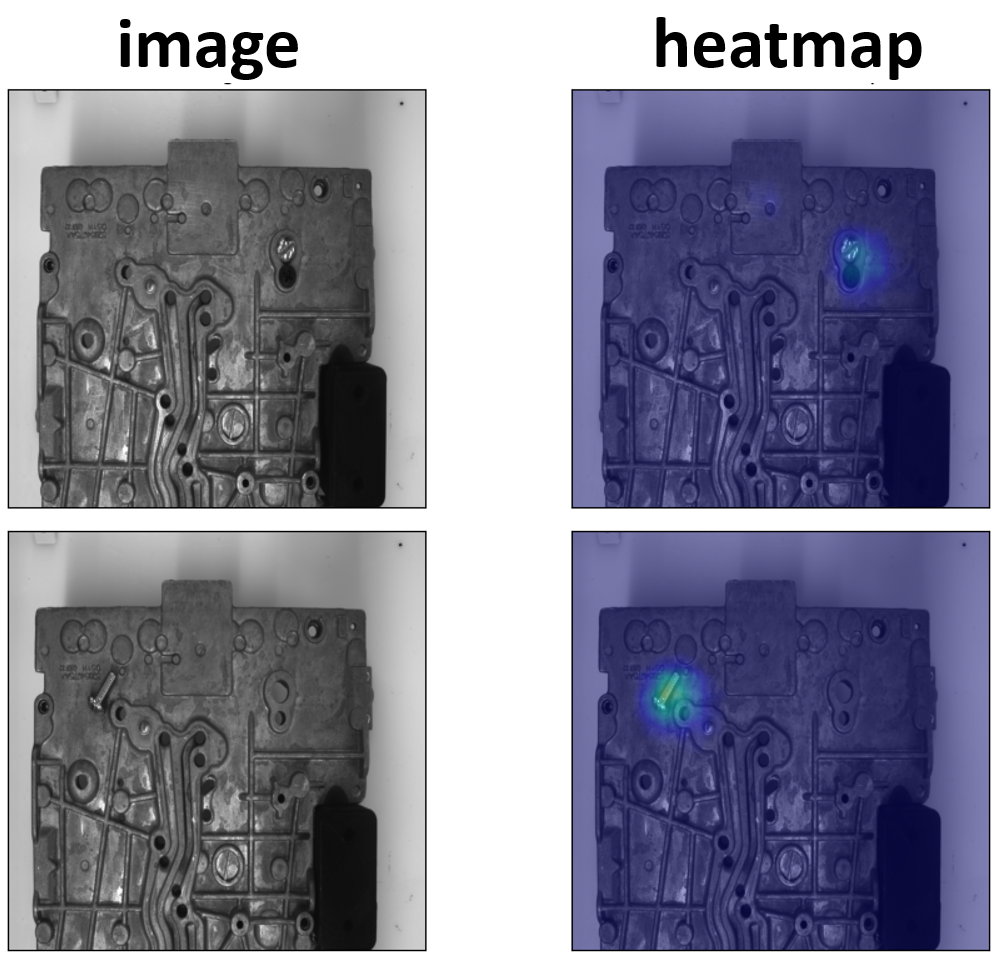} %
\caption{Defect localization heatmaps for valve body images. Brighter green regions indicate areas with higher anomaly scores.}
\label{fig:valve_local}
\end{figure}

The result on Collected data is shown in Table \ref{table:confusion_matrix}.As shown all 34 normal samples were correctly identified. This is reflected in the True Positives (TP) count of 34 in the confusion matrix. Out of 36 bad samples, 3 were incorrectly identified as normal. This is shown as the False Positives (FP) count of 3. The overall accuracy of the model is 95.71\%. This indicates a high level of correctness in the model's predictions across all samples.The F1 Score is 95.77\%, which suggests a balanced performance between precision and recall. This is particularly valuable in scenarios where both false positives and false negatives are equally costly.
Visualization of per-pixel anomaly heatmaps overlaid on valve body images is shown in Figure \ref{fig:valve_local}.

\subsection{Ablation Study}
\begin{table}[h]
\centering
\begin{tabular}{cccccc}
\toprule[2pt]
Flow Steps & 1     & 2     & 3     & 4     & 5     \\ 
\midrule[1pt]
Im-AUROC & \textbf{99.28} & 98.89 & 98.73 & 98.57 & 98.48 \\ 
\bottomrule[2pt]
\end{tabular}
\caption{image level AUROC score of PatchFlow with different number of Flow steps}
\label{table:flowstep}
\end{table}

We conduct a comprehensive ablation study to evaluate the relationship between the number of flow steps and model performance. The results are shown in Table \ref{table:flowstep}. As the number of flow steps increases, the performance of PatchFlow decreases. 

\begin{table}[h]
\centering
\begin{tabular}{lccc}
\toprule[2pt]
Number of Scale                   & 3     & 2     & 1     \\ 
\midrule[1pt]
\multicolumn{1}{c}{Image AUROC} & \textbf{99.28} & 98.84 & 97.78 \\ 
\bottomrule[2pt]
\end{tabular}

\caption{Relation between number of scale and Image level AUROC score}
\label{table:scale}
\end{table}
Besides we test the influence of number of scale in scale pyramid.As Shown in Table \ref{table:scale},  the model utilizing a 3-scale image pyramid achieves the best performance. As the number of scales increases, the model is able to obtain more abstract and generalized information about the image by viewing a larger portion of the scene or object.

\section{Conclusions}
In this work, we have proposed a novel approach for automated anomaly detection in industrial product images. Our method combines local neighbor-aware patch features with a normalizing flow model. A key contribution is the introduction of an adapter module, which bridges the gap between a generic pretrained feature extractor and specialized industrial product images. Additionally, we incorporated a bottleneck structure to reduce computational complexity while preserving the mapping capabilities of the flow architecture.

Through comprehensive empirical evaluation, our model achieves new state-of-the-art performance for image-level anomaly detection on the challenging MVTec AD dataset\cite{8954181} , reducing the error rate by 20\% compared to prior SOTA\cite{roth2021total} and attaining an image level AUROC score of 99.28\%  and VisA Dataset\cite{zou2022}, reducing the error rate by 28.2\% compared with \cite{roth2021total} and attaining an image level AUROC score of 96.48\%. PatchFlow's performance on both datasets, as highlighted by its leading AUROC score, represents a notable advancement in the field of anomaly detection, setting a new benchmark for future models and applications. Further experiments on a proprietary die casting dataset yield a defect detection accuracy of 95.77\%, without requiring any anomalous training data.

These results highlight the potential of leveraging computer vision and deep learning to enhance automated visual inspection for critical manufacturing techniques such as die casting. Our method illustrates a viable path to improving quality control and reducing waste in industrial production environments. This work constitutes meaningful progress, and could provide a strong foundation for future research and adoption in real-world inspection systems.


\clearpage
\bibliographystyle{plainnat}
\bibliography{ref}

@misc{zhang2023prototypical,
      title={Prototypical Residual Networks for Anomaly Detection and Localization}, 
      author={Hui Zhang and Zuxuan Wu and Zheng Wang and Zhineng Chen and Yu-Gang Jiang},
      year={2023},
      eprint={2212.02031},
      archivePrefix={arXiv},
      primaryClass={cs.CV}
}

@misc{lee2022cfa,
      title={CFA: Coupled-hypersphere-based Feature Adaptation for Target-Oriented Anomaly Localization}, 
      author={Sungwook Lee and Seunghyun Lee and Byung Cheol Song},
      year={2022},
      eprint={2206.04325},
      archivePrefix={arXiv},
      primaryClass={cs.CV}
}

@InProceedings{Tien_2023_CVPR,
    author    = {Tien, Tran Dinh and Nguyen, Anh Tuan and Tran, Nguyen Hoang and Huy, Ta Duc and Duong, Soan T.M. and Nguyen, Chanh D. Tr. and Truong, Steven Q. H.},
    title     = {Revisiting Reverse Distillation for Anomaly Detection},
    booktitle = {Proceedings of the IEEE/CVF Conference on Computer Vision and Pattern Recognition (CVPR)},
    month     = {June},
    year      = {2023},
    pages     = {24511-24520}
}

@INPROCEEDINGS{8954181,
  author={Bergmann, Paul and Fauser, Michael and Sattlegger, David and Steger, Carsten},
  booktitle={2019 IEEE/CVF Conference on Computer Vision and Pattern Recognition (CVPR)}, 
  title={MVTec AD — A Comprehensive Real-World Dataset for Unsupervised Anomaly Detection}, 
  year={2019},
  volume={},
  number={},
  pages={9584-9592},
  doi={10.1109/CVPR.20 19.00982}}

@inproceedings{goodfellow2014generative,
  title={Generative adversarial nets},
  author={Goodfellow, Ian and Pouget-Abadie, Jean and Mirza, Mehdi and Xu, Bing and Warde-Farley, David and Ozair, Sherjil and Courville, Aaron and Bengio, Yoshua},
  booktitle={Advances in neural information processing systems},
  pages={2672--2680},
  year={2014}
}

@misc{kingma2022autoencoding,
      title={Auto-Encoding Variational Bayes}, 
      author={Diederik P Kingma and Max Welling},
      year={2022},
      eprint={1312.6114},
      archivePrefix={arXiv},
      primaryClass={stat.ML}
}

@misc{dinh2017density,
      title={Density estimation using Real NVP}, 
      author={Laurent Dinh and Jascha Sohl-Dickstein and Samy Bengio},
      year={2017},
      eprint={1605.08803},
      archivePrefix={arXiv},
      primaryClass={cs.LG}
}

@misc{rezende2016variational,
      title={Variational Inference with Normalizing Flows}, 
      author={Danilo Jimenez Rezende and Shakir Mohamed},
      year={2016},
      eprint={1505.05770},
      archivePrefix={arXiv},
      primaryClass={stat.ML}
}

@misc{roth2021total,
      title={Towards Total Recall in Industrial Anomaly Detection},
      author={Karsten Roth and Latha Pemula and Joaquin Zepeda and Bernhard Schölkopf and Thomas Brox and Peter Gehler},
      year={2021},
      eprint={2106.08265},
      archivePrefix={arXiv},
      primaryClass={cs.CV}
}

@article{DBLP:journals/corr/abs-2011-08785,
  author       = {Thomas Defard and
                  Aleksandr Setkov and
                  Angelique Loesch and
                  Romaric Audigier},
  title        = {PaDiM: a Patch Distribution Modeling Framework for Anomaly Detection
                  and Localization},
  journal      = {CoRR},
  volume       = {abs/2011.08785},
  year         = {2020},
  url          = {https://arxiv.org/abs/2011.08785},
  eprinttype    = {arXiv},
  eprint       = {2011.08785},
  bibsource    = {dblp computer science bibliography, https://dblp.org}
}

@misc{yu2021fastflow,
      title={FastFlow: Unsupervised Anomaly Detection and Localization via 2D Normalizing Flows}, 
      author={Jiawei Yu and Ye Zheng and Xiang Wang and Wei Li and Yushuang Wu and Rui Zhao and Liwei Wu},
      year={2021},
      eprint={2111.07677},
      archivePrefix={arXiv},
      primaryClass={cs.CV}
}

@inproceedings{Gudovskiy_2022_WACV,
    author    = {Gudovskiy, Denis and Ishizaka, Shun and Kozuka, Kazuki},
    title     = {{CFLOW-AD}: Real-Time Unsupervised Anomaly Detection With Localization via Conditional Normalizing Flows},
    booktitle = {Proceedings of the IEEE/CVF Winter Conference on Applications of Computer Vision (WACV)},
    month     = {January},
    year      = {2022},
    pages     = {98-107}
}

@misc{ardizzone2019guided,
      title={Guided Image Generation with Conditional Invertible Neural Networks}, 
      author={Lynton Ardizzone and Carsten Lüth and Jakob Kruse and Carsten Rother and Ullrich Köthe},
      year={2019},
      eprint={1907.02392},
      archivePrefix={arXiv},
      primaryClass={cs.CV}
}

@inproceedings { RudWan2021,
author = {Marco Rudolph and Bastian Wandt and Bodo Rosenhahn},
title = {Same Same But DifferNet: Semi-Supervised Defect Detection with Normalizing Flows},
booktitle = {Winter Conference on Applications of Computer Vision (WACV)},
year = {2021},
month = jan
}

@inproceedings { RudWeh2022,
author = {Marco Rudolph and Tom Wehrbein and Bodo Rosenhahn and Bastian Wandt},
title = {Fully Convolutional Cross-Scale-Flows for Image-based Defect Detection},
booktitle = {Winter Conference on Applications of Computer Vision (WACV)},
year = {2022},
url = {arxiv},
month = jan
}

@misc{touvron2021training,
      title={Training data-efficient image transformers \& distillation through attention}, 
      author={Hugo Touvron and Matthieu Cord and Matthijs Douze and Francisco Massa and Alexandre Sablayrolles and Hervé Jégou},
      year={2021},
      eprint={2012.12877},
      archivePrefix={arXiv},
      primaryClass={cs.CV}
}

@misc{touvron2021going,
      title={Going deeper with Image Transformers}, 
      author={Hugo Touvron and Matthieu Cord and Alexandre Sablayrolles and Gabriel Synnaeve and Hervé Jégou},
      year={2021},
      eprint={2103.17239},
      archivePrefix={arXiv},
      primaryClass={cs.CV}
}

@INPROCEEDINGS{deng2009imagenet,
  author={Deng, Jia and Dong, Wei and Socher, Richard and Li, Li-Jia and Kai Li and Li Fei-Fei},
  booktitle={2009 IEEE Conference on Computer Vision and Pattern Recognition}, 
  title={ImageNet: A large-scale hierarchical image database}, 
  year={2009},
  volume={},
  number={},
  pages={248-255},
  doi={10.1109/CVPR.2009.5206848}}

@misc{tan2020efficientnet,
      title={EfficientNet: Rethinking Model Scaling for Convolutional Neural Networks}, 
      author={Mingxing Tan and Quoc V. Le},
      year={2020},
      eprint={1905.11946},
      archivePrefix={arXiv},
      primaryClass={cs.LG}
}

@misc{he2015deep,
      title={Deep Residual Learning for Image Recognition}, 
      author={Kaiming He and Xiangyu Zhang and Shaoqing Ren and Jian Sun},
      year={2015},
      eprint={1512.03385},
      archivePrefix={arXiv},
      primaryClass={cs.CV}
}

@misc{ronneberger2015unet,
      title={U-Net: Convolutional Networks for Biomedical Image Segmentation}, 
      author={Olaf Ronneberger and Philipp Fischer and Thomas Brox},
      year={2015},
      eprint={1505.04597},
      archivePrefix={arXiv},
      primaryClass={cs.CV}
}

@misc{sohldickstein2015deep,
      title={Deep Unsupervised Learning using Nonequilibrium Thermodynamics}, 
      author={Jascha Sohl-Dickstein and Eric A. Weiss and Niru Maheswaranathan and Surya Ganguli},
      year={2015},
      eprint={1503.03585},
      archivePrefix={arXiv},
      primaryClass={cs.LG}
}

@article{10.1145/3065386,
author = {Krizhevsky, Alex and Sutskever, Ilya and Hinton, Geoffrey E.},
title = {ImageNet classification with deep convolutional neural networks},
year = {2017},
issue_date = {June 2017},
publisher = {Association for Computing Machinery},
address = {New York, NY, USA},
volume = {60},
number = {6},
issn = {0001-0782},
url = {https://doi.org/10.1145/3065386},
doi = {10.1145/3065386},
journal = {Commun. ACM},
month = may,
pages = {84–90},
numpages = {7}
}

@article{aivision,
    title        = {A.I. Vision System for Automated Casting Quality Inspection},
    author       = {Boxiang Zhang and Baijian Yang and Corey Vian and Xiaoming Wang},
    year         = 2023,
    month        = {November},
    journal      = {Die Casting Engineer},
    volume       = {},
    number       = {},
    pages        = {10-16},
}

@article{sai2017critical,
  title={A critical review on casting types and defects},
  author={Sai, T Venkat and Vinod, T and Sowmya, Gunda},
  journal={Engineering and Technology},
  volume={3},
  number={2},
  pages={463--468},
  year={2017}
}

@misc{zagoruyko2017wide,
      title={Wide Residual Networks}, 
      author={Sergey Zagoruyko and Nikos Komodakis},
      year={2017},
      eprint={1605.07146},
      archivePrefix={arXiv},
      primaryClass={cs.CV}
}

@misc{li_cutpaste_2021,
      title={CutPaste: Self-Supervised Learning for Anomaly Detection and Localization}, 
      author={Chun-Liang Li and Kihyuk Sohn and Jinsung Yoon and Tomas Pfister},
      year={2021},
      eprint={2104.04015},
      archivePrefix={arXiv},
      primaryClass={cs.CV},
      url={https://arxiv.org/abs/2104.04015}, 
}

@InProceedings{Zavrtanik_2021_ICCV,
    author    = {Zavrtanik, Vitjan and Kristan, Matej and Skocaj, Danijel},
    title     = {DRAEM - A Discriminatively Trained Reconstruction Embedding for Surface Anomaly Detection},
    booktitle = {Proceedings of the IEEE/CVF International Conference on Computer Vision (ICCV)},
    month     = {October},
    year      = {2021},
    pages     = {8330-8339}
}

@InProceedings{Liu_2023_CVPR,
    author    = {Liu, Zhikang and Zhou, Yiming and Xu, Yuansheng and Wang, Zilei},
    title     = {SimpleNet: A Simple Network for Image Anomaly Detection and Localization},
    booktitle = {Proceedings of the IEEE/CVF Conference on Computer Vision and Pattern Recognition (CVPR)},
    month     = {June},
    year      = {2023},
    pages     = {20402-20411}
}

@InProceedings{zou2022,
author="Zou, Yang
and Jeong, Jongheon
and Pemula, Latha
and Zhang, Dongqing
and Dabeer, Onkar",
editor="Avidan, Shai
and Brostow, Gabriel
and Ciss{\'e}, Moustapha
and Farinella, Giovanni Maria
and Hassner, Tal",
title="SPot-the-Difference Self-supervised Pre-training for Anomaly Detection and Segmentation",
booktitle="Computer Vision -- ECCV 2022",
year="2022",
publisher="Springer Nature Switzerland",
address="Cham",
pages="392--408",
isbn="978-3-031-20056-4"
}
\end{document}